\documentclass[10pt,twocolumn,letterpaper]{article}

\usepackage{cvpr}
\usepackage{times}
\usepackage{epsfig}
\usepackage{graphicx}
\usepackage{amsmath}
\usepackage{amssymb}
\usepackage{booktabs}

\usepackage[pagebackref=true,breaklinks=true,colorlinks,bookmarks=false]{hyperref}

\begin{document}

\title{Is Synthetic Dataset Reliable for Benchmarking Generalizable Person Re-Identification?}

\author{Cuicui Kang\\
Mohamed bin Zayed University of Artificial Intelligence\\
Masdar City, Abu Dhabi, UAE\\
{\tt\small cuicui.kang@mbzuai.ac.ae}
}

\maketitle
\thispagestyle{empty}

\begin{abstract}

Recent studies show that models trained on synthetic datasets are able to achieve better generalizable person re-identification (GPReID) performance than that trained on public real-world datasets. On the other hand, due to the limitations of real-world person ReID datasets, it would also be important and interesting to use large-scale synthetic datasets as test sets to benchmark person ReID algorithms. Yet this raises a critical question: is synthetic dataset reliable for benchmarking generalizable person re-identification? In the literature there is no evidence showing this. To address this, we design a method called Pairwise Ranking Analysis (PRA) to quantitatively measure the ranking similarity and perform the statistical test of identical distributions. Specifically, we employ Kendall rank correlation coefficients to evaluate pairwise similarity values between algorithm rankings on different datasets. Then, a non-parametric two-sample Kolmogorov-Smirnov (KS) test is performed for the judgement of whether algorithm ranking correlations between synthetic and real-world datasets and those only between real-world datasets lie in identical distributions. We conduct comprehensive experiments, with ten representative algorithms, three popular real-world person ReID datasets, and three recently released large-scale synthetic datasets. Through the designed pairwise ranking analysis and comprehensive evaluations, we conclude that a recent large-scale synthetic dataset ClonedPerson can be reliably used to benchmark GPReID, statistically the same as real-world datasets. Therefore, this study guarantees the usage of synthetic datasets for both source training set and target testing set, with completely no privacy concerns from real-world surveillance data. Besides, the study in this paper might also inspire future designs of synthetic datasets.

\end{abstract}

\section{Introduction}

Person re-identification (ReID) aims at retrieving the same person as query images from a large pool of gallery images collected from surveillance videos. It has important values in both computer vision research as well as industry applications. Accordingly, person re-identification techniques have been largely advanced during the past decade, especially by deep learning \cite{gong2014person,WU2019neuro,ye2022survey}.
 
However, publicly available real-world person re-identification datasets are usually limited in scale for deep learning. For example, one of the largest and popular public dataset in surveillance, MSMT17 \cite{wei2018person}, has only 4,101 identities. Collecting even larger datasets is difficult, due to the expensive cost in labeling the same identity in different videos, as well as an important concern on the privacy issue for collecting surveillance data. As a result, synthetic datasets gain more and more attentions in person re-identification research \cite{wang20rand, zhang2021unrealperson, Wang-2022-Clonedperson}. 

Especially, recent studies \cite{wang20rand, zhang2021unrealperson, Wang-2022-Clonedperson} show that models trained on synthetic datasets are able to achieve better generalizable person re-identification (GPReID) results than that trained on public real-world datasets. As a result, it appears that it is promising to replace real-world person ReID datasets by synthetic datasets, with the advantages of better generalization performance, large scalability, and no privacy concern. However, this is only partially true, because in the literature it is only proved that synthetic dataset can be a better training source, but there is no evidence showing that it can be reliably used to benchmark GPReID algorithms as well. 
For example, in recent studies \cite{wang20rand, zhang2021unrealperson, Wang-2022-Clonedperson}, all synthetic datasets are used for training data, while real-world datasets are used as test data for benchmark evaluation of GPReID, except that though ClonedPerson \cite{Wang-2022-Clonedperson} has partitioned a test set, the limited evaluations with only one example algorithm for each task are not adequate to draw any conclusion on whether the synthetic dataset is reliable to benchmark GPReID algorithms.

However, large-scale benchmarking dataset is also important for person re-identification, especially for cross-dataset evaluation in GPReID where both source training dataset and target evaluation dataset are required. Furthermore, due to the limitations of real-world person ReID datasets discussed above, it would be also important and interesting to use large-scale synthetic person ReID datasets as target evaluation dataset for benchmarking, instead of considering enlarging expensive real-world ones. While existing studies have proved that synthetic datasets can be good training source, if it can be further proved that they are also good as test data for performance evaluation, we will be able to completely remove the dependency on privacy-sensitive real-world surveillance data for person re-identification research. Yet this raises a critical question: \textbf{\textit{is synthetic dataset reliable for benchmarking generalizable person re-identification?}} In another word, will algorithm rankings evaluated on synthetic datasets similar to that evaluated on real-world datasets? There is an intuitive doubt on this question, because the characteristics of synthetic datasets are quite different from real-world datasets.

Consequently, to address the above question, in this work, we conduct comprehensive experiments, with ten representative algorithms, three popular real-world person ReID datasets (CUHK03 \cite{li2014deepreid}, Market-1501 \cite{zheng2015scalable}, and MSMT17 \cite{wei2018person}), and three large-scale synthetic datasets (RandPerson \cite{wang20rand}, UnrealPerson \cite{zhang2021unrealperson}, and ClonedPerson \cite{Wang-2022-Clonedperson}). 
We design a number of cross-dataset evaluation experiments for GPReID accordingly. From the benchmarking results, we compare the relative performance among different algorithms, on both real-world and synthetic test datasets. Through qualitative observations, we find that algorithm rankings on the synthetic dataset are quite similar to those on the real-world datasets. This means that comparing algorithms on the synthetic dataset gives a strong agreement to comparing them on real-world datasets. 


More formally, we design a method called Pairwise Ranking Analysis (PRA) to statistically test if the usage of synthetic datasets would be the same as real-world datasets in benchmarking algorithms. Specifically, we employ Kendall rank correlation coefficients to evaluate pairwise similarity values between algorithm rankings on different datasets. Then, a non-parametric two-sample Kolmogorov-Smirnov (KS) test is performed for the judgement of whether algorithm ranking correlations between synthetic and real-world datasets and those only between real-world datasets lie in identical distributions. Through comprehensive evaluations, we conclude that the most recent large-scale synthetic dataset ClonedPerson can be reliably used to benchmark GPReID, statistically the same as real-world datasets. Therefore, this study guarantees the usage of synthetic datasets for both source training set and target testing set, with completely no privacy concerns from real-world surveillance data. Besides, the study in this paper might also inspire future designs of synthetic datasets.

\section{Related Works}



Person ReID techniques have been largely advanced in the past decade. Recently, generalizable person ReID has gained lots of attentions due to common domain shifts in application scenarios. We briefly reviewed ten representative person ReID algorithms used in this paper in Section \ref{sec:algo}, with five traditional methods, and five GPReID methods. For more comprehensive survey, please refer to \cite{gong2014person,WU2019neuro,ye2022survey}.

One naive idea to improve the generalization ability is expanding the training dataset. However, publicly available real-world datasets are mostly limited in scale due to the expensive and hard manually labeling work, as well as privacy concerns in collecting surveillance data. For example, the most popularly used person ReID datasets, CUHK03 \cite{li2014deepreid}, Market-1501 \cite{zheng2015scalable}, and MSMT17 \cite{wei2018person}, contain only 1,360, 1,501, and 4,101 identities, respectively. 

Thus, a new trending is to develop large-scale synthetic datasets to improve the domain generalization ability. 
Compared to real-world data, synthetic data can be easily expanded to large scale, and there is no privacy issue and no need of hard manually labeling. 
For early tries, Barbosa \etal \cite{barbosa2018looking} proposed the synthetic dataset SOMAset with 50 person models. Particularly, each model has 11 types of outfits. Bak \etal \cite{bak2018domain} used hand-crafted characters to generate data and produced the SyRI dataset with 100 subjects. Sun \etal \cite{sun2019dissecting} proposed the PersonX dataset with 1,266 characters, but it is still limited with hand-crafted 3D person models. Interestingly, PersonX randomly partitioned data into 410 identities for training and the rest 856 identities for testing, and evaluated three algorithms. With the experiments it concluded that the performance trend of the three algorithms is similar between PersonX and real-world datasets. However, this was for traditional single-source training and testing, but not cross-dataset evaluaiton in GPReID. What's more, it only validated three algorithms, and the conclusion was drawn based merely on rough observations. Recently, Wang \etal \cite{wang20rand} created a large-scale synthetic dataset RandPerson with 8,000 characters, by generating new-looking clothes models with random colors and patterns to replace UV maps of existing 3D clothes models. Then, inspired by the method of RandPerson, Zhang \etal \cite{zhang2021unrealperson} proposed the UnrealPerson dataset, with 3,000 virtual characters, by replacing UV maps with cropped clothes patches from real-world person images. More recently, Wang \etal \cite{Wang-2022-Clonedperson} developed the ClonedPerson dataset by systematically cloning outfits from real-world person images to virtual 3D characters with clear clothing textures. 

Furthermore, recent developments of \cite{wang20rand,zhang2021unrealperson,Wang-2022-Clonedperson} proved that models trained on synthetic datasets can generalize well on real-world datasets, showing a promising future of using synthetic datasets instead of real-world datasets for training to avoid privacy issues and expensive data labeling. 
Inspired by using synthetic datasets for large-scale training, we are wondering if synthetic datasets could be reliably used for benchmarking GPReID. Considering this, this work designs particular methods and comprehensive experiments to answer this question, which will advance further studies with synthetic datasets for even wider applications in the future.

\section{Method}

\subsection{Methodology Overview}



Cross-dataset evaluation is typically adopted to benchmark the performance of generalizable person re-identification \cite{yi2014deep, hu2014cross, liao2020interpretable}. That is, algorithms are required to be trained on a source training dataset, while evaluated on an independent target dataset to understand their generalizability. Traditionally, only real-world datasets have been considered as target datasets. This is the most interesting setting because it is intuitive to use such benchmarking results to guide real-world applications.

In this paper, we try to additionally include synthetic datasets as target datasets for benchmarking. Then, there will be a set of results on real-world target datasets, as well as a set of results on synthetic target datasets, for a set of algorithms to be evaluated. Accordingly, we rank algorithms within each set based on their performance. Then, to understand the reliability of benchmarking on synthetic datasets, we compare algorithm rankings on synthetic datasets to that on real-world datasets. Ideally, we expect that they would be the same or very similar. However, due to dataset bias and other influential factors, this is hard to achieve.

Therefore, to understand how the algorithm rankings are similar between evaluations on real-world datasets and synthetic datasets, we perform qualitative analysis and design a method called Pairwise Ranking Analysis (PRA) to quantitatively measure the ranking similarity and perform the statistical test of identical distributions. For qualitative analysis, we plot all algorithms' performance curves to understand their consistency in order. For quantitative analysis by PRA, first, we employ Kendall rank correlation coefficients to evaluate pairwise similarity values between algorithm rankings on different datasets. Second, ranking similarity values between results evaluated on real-world datasets are used as reference values and form a reference distribution, while those between a synthetic dataset and other real-world datasets are treated as inspecting values and form an inspecting distribution. Then, a hypothesis test is constructed for the judgement of whether the two distributions are identical. The non-parametric two-sample Kolmogorov-Smirnov (KS) test is employed for the test, since we do not know the type of the distributions. Finally, if the null hypothesis is accepted given a confidence level of 95\%, we conclude that the synthetic dataset is as reliable as real-world datasets to benchmark generalizable person re-identification algorithms, or otherwise it is not reliable.

\subsection{Pairwise Ranking Analysis}

As introduced above, the pairwise ranking analysis is designed to measure the similarity values between pairs of algorithm rankings and perform the statistical test of identical distributions. To achieve this, the Kendall rank correlation coefficient is imported to evaluate the relationships between two algorithm rankings, while the two-sample KS test is employed for the identical distribution test.

\textbf{Kendall's $\tau$:} The Kendall correlation coefficient $\tau$ is a statistic to measure the correlation between two ordinal variables. The main idea of Kendall's $\tau$ is to use the numbers of concordant pairs and discordant pairs of two variables to calculate the correlation. Let $\mathbf{x}$ and $\mathbf{y}$ be two random variables, whose observations are $\{x_i\}$ and $\{y_i\}$ for $i \in \{1,\dots,N\}$, respectively. $\forall i \neq j$, if both $x_i < x_j$ and $y_i < y_j$ hold, or both $x_i > x_j$ and $y_i > y_j$ hold, the pair of observations ($x_i$, $y_i$) and ($x_j$, $y_j$) are said to be concordant. Otherwise, they are discordant. Then, the Kendall correlation coefficient with tie adjustments is defined as \cite{kendall1945}:
\begin{equation}
  \tau = \frac{n_{c} - n_{d}}{\sqrt{(n_0 - n_1 )(n_0 - n_2 )}} ,
\end{equation}
where $n_c$ is the number of concordant pairs, $n_d$ is the number of discordant pairs, \( n_0 = N(N-1)/2 \), \(n_1 = \sum_{i}t_i (t_i -1)/2\), and \( n_2 =\sum_j u_j (u_j -1)/2 \), with $t_i$ being the number of tied values in the $i_{th}$ group of ties for $\mathbf{x}$ only, and $u_j$ being the number of tied values in the $j_{th}$ group of ties for $\mathbf{y}$ only. It is also called as Kendall's $\tau_b$ statistic, which makes adjustments for ties \cite{kendall1945}.

The value range of \(\tau\) is in [-1, 1]. When $\tau$ = 1, it means that the two variables have identical rank correlation. When $\tau$ = -1, it means that the two variables have completely opposite rank correlation. When $\tau$ = 0, the two variables are not correlated.

In our evaluation, we have a set of $N$ algorithms, being evaluated on a set of target datasets, both real-world and synthetic. Then, for each target dataset, we will have a set of $N$ results for a certain performance metric. Accordingly, we compute Kendall's $\tau$ between the sets of $N$ results on any pair of target datasets.

\textbf{Kolmogorov-Smirnov Test:} With Kendall's $\tau$ values computed, we further divide the values into two groups: Group A with $\tau$ values computed only between real-world target datasets, and Group B with $\tau$ values computed between real-world and synthetic target datasets. Observations in Group A with $n$ values form a reference distribution $F_n$, which describes how similarity values of algorithm rankings between real-world target datasets vary. On the other hand, Observations in Group B with $m$ values form an inspecting distribution $G_m$. If $G_m$ is identical to $F_n$, we may conclude that the inspecting synthetic target dataset has no statistical difference to real-world datasets in benchmarking algorithms.

Accordingly, we construct a hypothesis test, with the null hypothesis being $H_0$: $G_m$ is identical to $F_n$, and its alternative hypothesis being $H_1$: $G_m$ is different to $F_n$. Since we do not know the distribution type of either $F_n$ or $G_m$, we apply the non-parametric two-sample Kolmogorov-Smirnov test, which is commonly used to test whether two distributions are identical \cite{NAAMAN2021109088}. Specifically, with the two empirical distribution functions $F_n$ and $G_m$, the Kolmogorov–Smirnov statistic is

\begin{equation}
    D_{n, m} = \sup _{x}|F_n(x) - G_m(x)|,
\end{equation}
where $\sup$ is the supremum function. Then the null hypothesis is rejected at level $\alpha$ if

\begin{equation} \label{eq:D-thr}
    D_{n,m} > {\sqrt {-\ln \left({\tfrac {\alpha }{2}}\right)\cdot {\tfrac {1+{\tfrac {m}{n}}}{2m}}}}.
\end{equation}

Or alternatively, a p-value is computed according to the KS distribution at $D_{n,m}$. If the p-value is lower than the significance level $\alpha$, the null hypothesis will be rejected. Otherwise, it will be accepted, that is, we will consider that $G_m$ is identical to $F_n$, and thus the inspecting synthetic dataset has no statistical difference to real-world datasets in benchmarking algorithms.

\subsection{Person Re-Identification Algorithms}\label{sec:algo}


We select ten person re-identification algorithms for this study, with five traditional methods and the other five from recent developments in generalizable person ReID.

\textbf{Traditional Methods:}
\textbf{(1) PCB:} In order to learn discriminative part informed features for person ReID, Sun \etal \cite{sun2018pcb} proposed the Part-based Convolutional Baseline (PCB), which used a simple uniform partition strategy and aggregates part features into a descriptor for more discriminative features. 
\textbf{(2) MLFN:} The Multi-Level Factorisation Net (MLFN) architecture \cite{chang18mlfn} targets at learning discriminative and view-invariant visual factors of identities at multiple semantic levels. It is composed of multiple stacked blocks containing multiple factor modules and a factor selection module that factorise the visual appearance of identities into latent discriminative factors at different levels. 
\textbf{(3) MGN:} The Multiple Granularity Network (MGN) \cite{wang2018learning} is developed with an end-to-end learning strategy based on ResNet50 \cite{he2016deep}, which is connected to three branches containing one branch for global feature learning and two branches for local feature learning with multi-level granularity. 
\textbf{(4) OSNet:} Zhou \etal \cite{Zhou2019-OSNet} proposed the OSNet, a light weight but efficient network to learn Omni Scale representations for person re-identification task. The basic block in OSNet is a residual block composed of multiple convolutional streams and unified aggregation gate to dynamically fuse multi-scale features with input-dependent channel-wise weights. 
\textbf{(5) AGW:} AGW \cite{ye2022survey} is designed on top of BagOfTricks \cite{Luo19BOT} for improved performance, with additional non-local attention block, generalized mean pooling and weighted regularized triplet loss.

\textbf{GPReID Methods:}
\textbf{(1) ResNet50-mid:} Yu \etal \cite{yu2017devil} proposed to use mid-level features in a deep neural architecture and proved their advantage for cross-domain instance matching. 
\textbf{(2) OSNet-IBN:} Based on OSNet, OSNet-IBN \cite{Zhou2019-OSNet} was proposed with Instance Normalization (IN) \cite{pan2018two} to improve the generalisation performance.
\textbf{(3) OSNet-AIN:} OSNet-AIN \cite{zhou2021osnet} is developed based on the combination of OSNet and IN layers for better generalization ability of person re-identification. It employs an efficient differentiable architecture search algorithm to determine optimal placements of the IN layers in OSNet.
\textbf{(4) QAConv-GS:} Liao and Shao \cite{liao2020interpretable} proposed the Query Adaptive Convolution (QAConv) for generalizable person re-identification, which uses explicit deep feature matching and proves that matching between pairs of deep feature maps is effective and generalizable for person ReID. They further proposed the improved version QAConv-GS with Graph Sampling (GS) for more efficient learning \cite{Liao-2021-QAConv-GS}. We adopt the improved version QAConv-GS in this paper.
\textbf{(5) TransMatcher:} Liao and Shao \cite{liao2021transmatcher} further proposed TransMatcher to use Transformers for generalizable person re-identification. It performs query-gallery cross-attention matching in a simplified decoder where the full attention implementation is replaced with query-key similarity computation. This design shows further improvements to GPReID.

\subsection{Evaluation Pipeline}
With the above ten algorithms, for each experiment on one target dataset, we will obtain a vector of ten elements for a certain performance metric. Then, the Kendall's $\tau$ will work on these size-10 vectors, and compute $\tau$ values for each pair of these size-10 vectors. Finally, the KS test will work on the distributions of the $\tau$ values. Specifically, these $\tau$ values will be divided into two groups: one group is associated with only real-world datasets, while the other group contains $\tau$ values between real-world and synthetic datasets. The KS test will determine if the distributions of the $\tau$ values in the two groups are identical or not, and so conclude if any synthetic dataset can be used for benchmarking with no statistical difference to real-world datasets.

\section{Experiments}

\subsection{Datasets}

\textbf{Real-World Datasets:} Three public and most popular large-scale real-world person re-identification datasets, CUHK03 \cite{li2014deepreid}, Market-1501 \cite{zheng2015scalable}, and MSMT17 \cite{wei2018person}, are used in the experiments. 
The CUHK03 dataset is built with 1,360 pedestrians and 13,164 images. The train/test data split in \cite{zhong2017re} is followed for person Re-ID task, which is called new protocol (NP), where 767 and 700 subjects are used for training and testing, respectively. Besides, the “detected” subset is used instead of the “labeled”, which contains 7,365 images for training, 5,332 images for gallery and 1,400 images for query. The Market-1501 dataset was captured from six cameras and consists of 32,668 images with 1,501 identities. The official data split in \cite{zheng2015scalable} is used, with 750 identities for training and the remaining 751 identities for testing. The Multi-Scene Multi-Time person ReID dataset, namely MSMT17, contains 4,101 identities and 126,441 images captured from 15 cameras with both indoor and outdoor scenes \cite{wei2018person}. We followed the official data split protocol, which contains 32,621 images from 1,041 identities for training set, and the remaining images from 3,060 identities as the testing set. 

\textbf{Synthetic Datasets:} Three synthetic datasets are used in the experiments, namely RandPerson \cite{wang20rand}, UnrealPerson \cite{zhang2021unrealperson} and ClonedPerson \cite{Wang-2022-Clonedperson}.
The RandPerson dataset is the first synthetic person re-identification dataset used to improve generalization ability for person re-identification task \cite{wang20rand}. The dataset contains 1,801,816 synthesized images of 8,000 identities, which are generated by Unity3D with surveillance environments simulation. A subset including 132,145 images of the 8,000 identities suggested in \cite{wang20rand} is used in the experiments. The UnrealPerson dataset contains 6,799 identities. A random subset suggested in \cite{zhang2021unrealperson} containing 120,000 images from 3,000 identities is used in the experiments. Note that both RandPerson and UnrealPerson have no test set partition, and so they are only used as training datasets following \cite{wang20rand,zhang2021unrealperson}. 
ClonedPerson is a newly released large-scale synthetic dataset for GPReID \cite{Wang-2022-Clonedperson}, which contains 5,621 identities and 887,766 images. In order to generate more realistic images rather than cartoon-like images, Wang \etal proposed to clone outfits from real-world person images to virtual 3D characters to make the identities look more similar to the real-world counterparts. The dataset is divided to a training set with 763,953 images from 4,826 characters and a testing set with 123,813 images from 795 characters. Thus, it is available for both training and testing for generalizable person re-identification.


\subsection{Experimental Settings}

Three open-source projects,  FastReID\footnote{https://github.com/JDAI-CV/fast-reid} \cite{he2020fastreid}, Torchreid\footnote{https://github.com/KaiyangZhou/deep-person-reid} \cite{torchreid}, and QAConv\footnote{https://github.com/ShengcaiLiao/QAConv} \cite{liao2020interpretable,Liao-2021-QAConv-GS} are used for the evaluation of the ten algorithms listed in Section \ref{sec:algo}. In the experiments, the default configuration is used for all the three projects, except for the combineall parameter, which is set to False for all tasks. Besides, if the methods in these projects are available for multiple basic CNN backbones, the ResNet50 with IBN \cite{pan2018two,jia2019frustratingly} is used for all available methods.

For the six datasets, if there are training and testing subset partitions, training is performed on the training subset only, while the testing subset is used for testing only. Except that for RandPerson and UnrealPerson, they are only used for training. Cross-dataset evaluation is performed for GPReID \cite{yi2014deep,hu2014cross}, with training on the training subset of one dataset, and testing on the testing subset of another dataset. All evaluations follow the single-query evaluation protocol. For the evaluation metrics, the Rank-1 (R1) accuracy and mean average precision (mAP) are used as the performance measurement criteria. 


\begin{table}
\centering
\begin{tabular}{|@{~}l@{}|r@{~~~}r|r@{~~~}r|r@{~~~}r|}
\hline
  & \multicolumn{4}{c|}{Real-World}&\multicolumn{2}{c|}{Synthetic} \\ \hline
Dataset  & \multicolumn{2}{@{}c@{}|}{MSMT17}&\multicolumn{2}{@{}c@{}|}{Market-1501} & \multicolumn{2}{@{}c@{}|}{ClonedPerson}  \\ \hline
Method & R1 & mAP & R1 & mAP & R1 & mAP \\  \hline 
PCB          &  6.1 & 1.6  & 37.9 & 17.1 & 10.9 & 0.8  \\ \hline  
MLFN            &  7.7 & 2.4  & 40.4 & 19.1 & 10.7 & 0.8  \\ \hline  
Resnet50mid     &  8.3 & 2.5  & 43.2 & 20.6 & 10.2 & 0.7  \\ \hline  
OSNet           & 11.0 & 3.4  & 47.4 & 22.7 & 15.3 & 1.1  \\ \hline  
OSNet-IBN       & 27.9 & 9.1  & 58.9 & 31.0 & 20.9 & 2.2  \\ \hline  
OSNet-AIN       & 27.5 & 8.9  & 58.5 & 30.1 & 21.3 & 2.2  \\ \hline  
AGW             & 16.1 & 5.2  & 57.5 & 32.8 & 21.8 & 2.1  \\ \hline
MGN             & 28.0 & 9.3  & 63.9 & 37.4 & 28.9 & 3.5  \\ \hline   
QAConv-GS       & 46.9 & 15.7 & 68.5 & 37.1 & 32.8 & 4.0  \\ \hline  
TransMatcher    & 46.8 & 15.7 & 70.0 & 40.5 & 41.4 & 5.3  \\  
\hline
\end{tabular}
\caption{Results with CUHK03 as training set.}
\label{tlb:cross-cuhk}
\end{table}

\begin{table}
\centering
\begin{tabular}{|@{~}l@{}|r@{~~~}r|r@{~~~}r|r@{~~~}r|}
\hline
  & \multicolumn{4}{c|}{Real-World}&\multicolumn{2}{c|}{Synthetic} \\ \hline
Dataset  & \multicolumn{2}{@{}c@{}|}{MSMT17}&\multicolumn{2}{@{}c@{}|}{CUHK03} & \multicolumn{2}{@{}c@{}|}{ClonedPerson}  \\ \hline
Method & R1 & mAP & R1 & mAP & R1 & mAP \\  \hline
PCB          &  7.2 & 2.1  &  5.2 & 5.4  & 13.5 & 1.2 \\ \hline    
MLFN            &  8.7 & 2.7  &  4.1 & 3.9  & 12.6 & 1.3 \\ \hline    
Resnet50mid     &  8.2 & 2.6  &  4.4 & 4.4  & 13.2 & 1.3 \\ \hline    
OSNet           & 12.1 & 4.0  &  7.2 & 7.7  & 20.9 & 2.1 \\ \hline    
OSNet-IBN       & 23.7 & 7.9  & 10.9 & 10.4 & 30.8 & 4.2 \\ \hline    
OSNet-AIN       & 24.7 & 8.3  & 11.0 & 10.4 & 31.7 & 4.5 \\ \hline    
AGW             & 15.8 & 5.6  & 11.0 & 10.8 & 28.1 & 3.2 \\ \hline     
MGN             & 32.4 & 12.0 & 19.6 & 19.3 & 39.2 & 6.2 \\ \hline        
QAConv-GS       & 47.6 & 17.7 & 18.4 & 17.7 & 41.2 & 6.8 \\ \hline    
TransMatcher    & 48.0 & 18.6 & 20.8 & 20.1 & 50.1 & 9.2 \\ 
\hline
\end{tabular}
\caption{Results with Market-1501 as training set.}
\label{tlb:cross-market}
\end{table}

\begin{table}
\centering
\begin{tabular}{|@{~}l@{}|r@{~~~}r|r@{~~~}r|r@{~~~}r|}
\hline
  & \multicolumn{4}{c|}{Real-World}&\multicolumn{2}{c|}{Synthetic} \\ \hline
Dataset  & \multicolumn{2}{@{}c@{}|}{Market-1501}&\multicolumn{2}{@{}c@{}|}{CUHK03} & \multicolumn{2}{@{}c@{}|}{ClonedPerson}  \\ \hline
Method & R1 & mAP & R1 & mAP & R1 & mAP \\  \hline 
PCB          & 40.9& 18.3&  7.3&  7.4& 15.9& 1.7  \\ \hline
MLFN            & 41.4& 19.0&  9.1&  9.3& 15.8& 1.8  \\ \hline
Resnet50mid     & 43.6& 20.7& 10.2& 10.7& 16.2& 1.8  \\ \hline
OSNet           & 58.0& 29.5& 14.3& 13.8& 26.9& 3.4  \\ \hline
OSNet-IBN       & 63.9& 32.6& 14.6& 14.0& 28.7& 3.9  \\ \hline
OSNet-AIN       & 63.7& 33.1& 15.6& 15.2& 29.4& 4.0  \\ \hline
AGW             & 63.2& 35.9& 17.1& 18.6& 31.2& 4.8  \\ \hline
MGN             & 73.8& 43.9& 21.8& 22.6& 39.2& 6.3  \\ \hline
QAConv-GS       & 78.4& 49.7& 19.6& 20.4& 44.1& 7.8  \\ \hline
TransMatcher    & 80.2& 52.1& 22.8& 22.4& 51.8& 9.0  \\ 
\hline
\end{tabular}
\caption{Results with MSMT17 as training set.}
\label{tlb:cross-msmt}
\end{table}

\begin{table*}[t]
\begin{minipage}[t]{\columnwidth}
\begin{tabular}{|@{~}l@{}|r@{~~}r@{~~}|r@{~~}r@{~~}|@{}r@{~~}r@{~~}|r@{~~}r@{~~}|}
\hline
  & \multicolumn{6}{c|}{Real-World}&\multicolumn{2}{c|}{Synthetic} \\ \hline
Dataset & \multicolumn{2}{@{}c@{}|}{MSMT17}&\multicolumn{2}{@{}c@{}|}{Market-1501} &\multicolumn{2}{@{}c@{}|}{CUHK03}& \multicolumn{2}{@{}c@{}|}{ClonedPerson}  \\ \hline
Method & R1 & mAP & R1 & mAP & R1 & mAP & R1 & mAP \\  \hline 
PCB          &  7.9 &  2.4 & 43.1 & 20.1 &  6.4 &  6.4 & 38.0 & 6.4    \\ \hline 
MLFN            &  6.1 &  1.8 & 37.4 & 16.6 &  5.3 &  5.6 & 35.3 & 5.8    \\ \hline 
Resnet50mid     &  5.7 &  1.7 & 36.6 & 16.4 &  4.8 &  5.0 & 34.3 & 5.6    \\ \hline 
OSNet           &  8.6 &  2.6 & 46.4 & 21.9 &  7.9 &  7.1 & 43.9 & 8.1    \\ \hline 
OSNet-IBN       & 15.3 &  4.7 & 50.2 & 24.7 &  9.6 &  8.7 & 49.6 & 12.0   \\ \hline 
OSNet-AIN       & 17.1 &  5.3 & 51.4 & 26.1 &  9.2 &  9.0 & 50.1 & 12.4   \\ \hline 
AGW             &  9.2 &  2.8 & 48.1 & 24.0 &  7.6 &  7.1 & 49.4 & 9.0    \\ \hline 
MGN             & 18.5 &  5.8 & 57.2 & 31.4 &  8.8 &  8.4 & 59.4 & 17.1   \\ \hline 
QAConv-GS       & 44.4 & 15.3 & 75.5 & 46.4 & 17.1 & 15.2 & 65.3 & 19.9   \\ \hline 
TransMatcher    & 45.2 & 16.2 & 77.2 & 48.8 & 19.2 & 17.8 & 67.8 & 22.1   \\       
\hline
\end{tabular}
\caption{Results with RandPerson as training set.}
\label{tlb:cross-rand}   
\end{minipage}
\hspace{24pt}
\begin{minipage}[t]{\columnwidth}
\centering
\begin{tabular}{|r@{~~}r@{~~}|r@{~~~}r@{~~}|r@{~~}r@{~~}|r@{~~}r@{~~}|}
\hline
 \multicolumn{6}{|c|}{Real-World}&\multicolumn{2}{c|}{Synthetic} \\ \hline
 \multicolumn{2}{|@{}c@{}|}{MSMT17}&\multicolumn{2}{@{}c@{}|}{Market-1501} &\multicolumn{2}{@{}c@{}|}{CUHK03}& \multicolumn{2}{@{}c@{}|}{ClonedPerson}  \\ \hline
 R1 & mAP & R1 & mAP & R1 & mAP & R1 & mAP \\  \hline
 11.9 &  3.8	 & 45.0 & 21.9 &  5.4 &  5.5 & 25.4 & 3.6    \\ \hline         
 11.6 &  3.4	 & 43.8 & 21.2 &  4.5 &  4.3 & 23.6 & 2.7    \\ \hline         
 9.9 &  3.0	 & 40.2 & 19.6 &  5.8 &  5.8 & 24.1 & 3.5    \\ \hline         
 14.7 &  4.5	 & 52.1 & 25.3 &  7.1 &  6.8 & 28.3 & 3.3    \\ \hline         
 24.3 &  8.0	 & 61.8 & 33.2 & 10.9 &  9.9 & 40.3 & 7.4    \\ \hline         
 25.2 &  8.2	 & 61.7 & 34.5 & 12.1 & 10.6 & 41.2 & 8.3    \\ \hline         
 19.0 &  6.4	 & 59.0 & 33.5 &  9.8 &  8.8 & 41.1 & 6.4    \\ \hline         
 27.2 &  9.2	 & 68.1 & 41.7 & 12.9 & 12.0 & 53.4 & 11.6   \\ \hline         
 52.5 & 20.0	 & 79.7 & 52.6 & 17.3 & 16.1 & 56.9 & 15.1   \\ \hline         
 52.0 & 21.3	 & 81.6 & 59.5 & 21.8 & 20.5 & 66.0 & 20.0   \\ 
\hline
\end{tabular}
\caption{Results with UnrealPerson as training set.}
\label{tlb:cross-unreal}   
\end{minipage}
\end{table*}


\begin{figure*}
\centering
  \begin{subfigure}[b]{0.19\textwidth}
  \includegraphics[width=\textwidth]{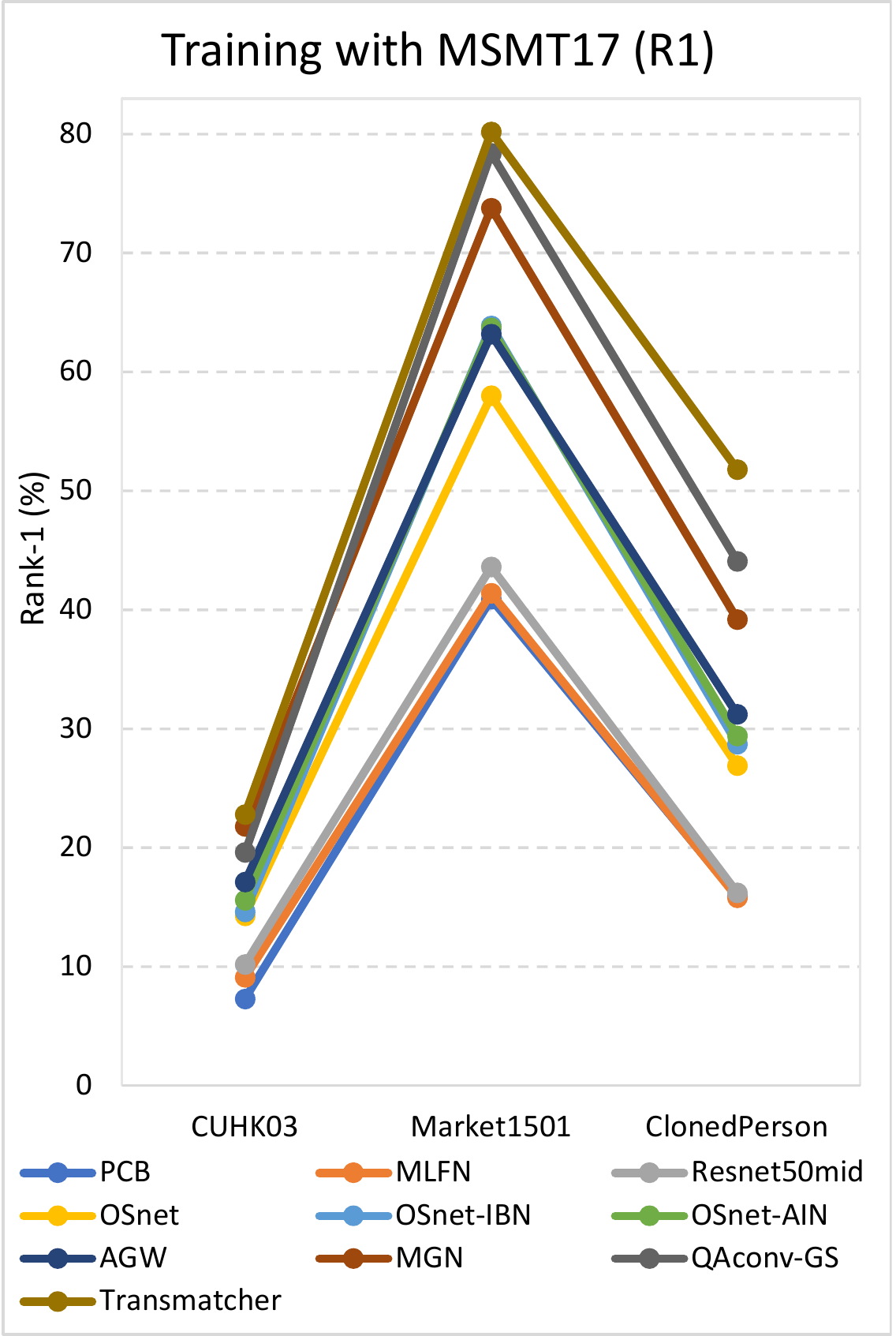}
  \caption{}\label{fig:msmt-r1}
  \end{subfigure}
  \begin{subfigure}[b]{0.184\textwidth}
  \includegraphics[width=\textwidth]{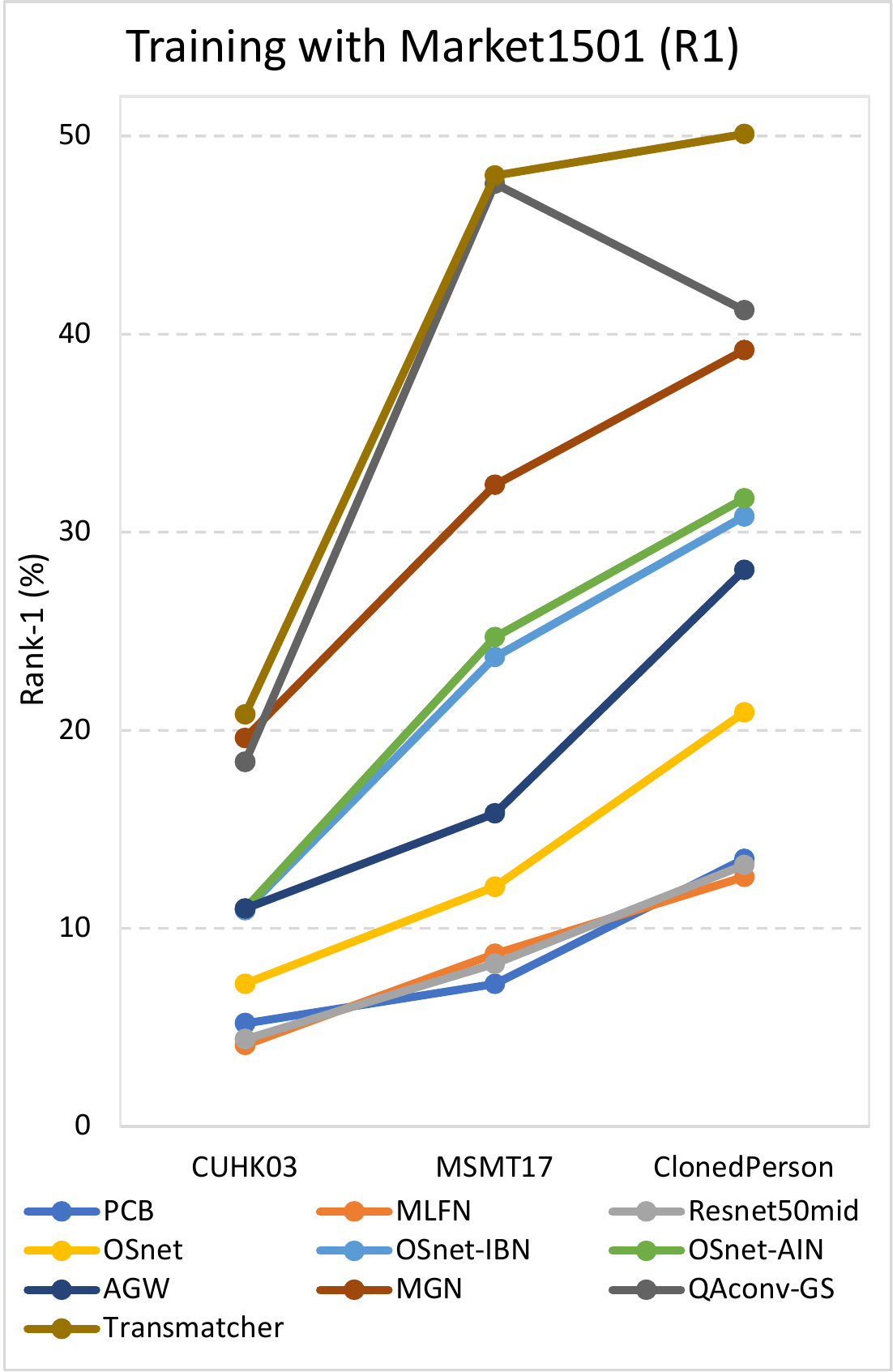}
  \caption{}\label{fig:mart-r1}
  \end{subfigure}
  \begin{subfigure}[b]{0.1855\textwidth}
  \includegraphics[width=\textwidth]{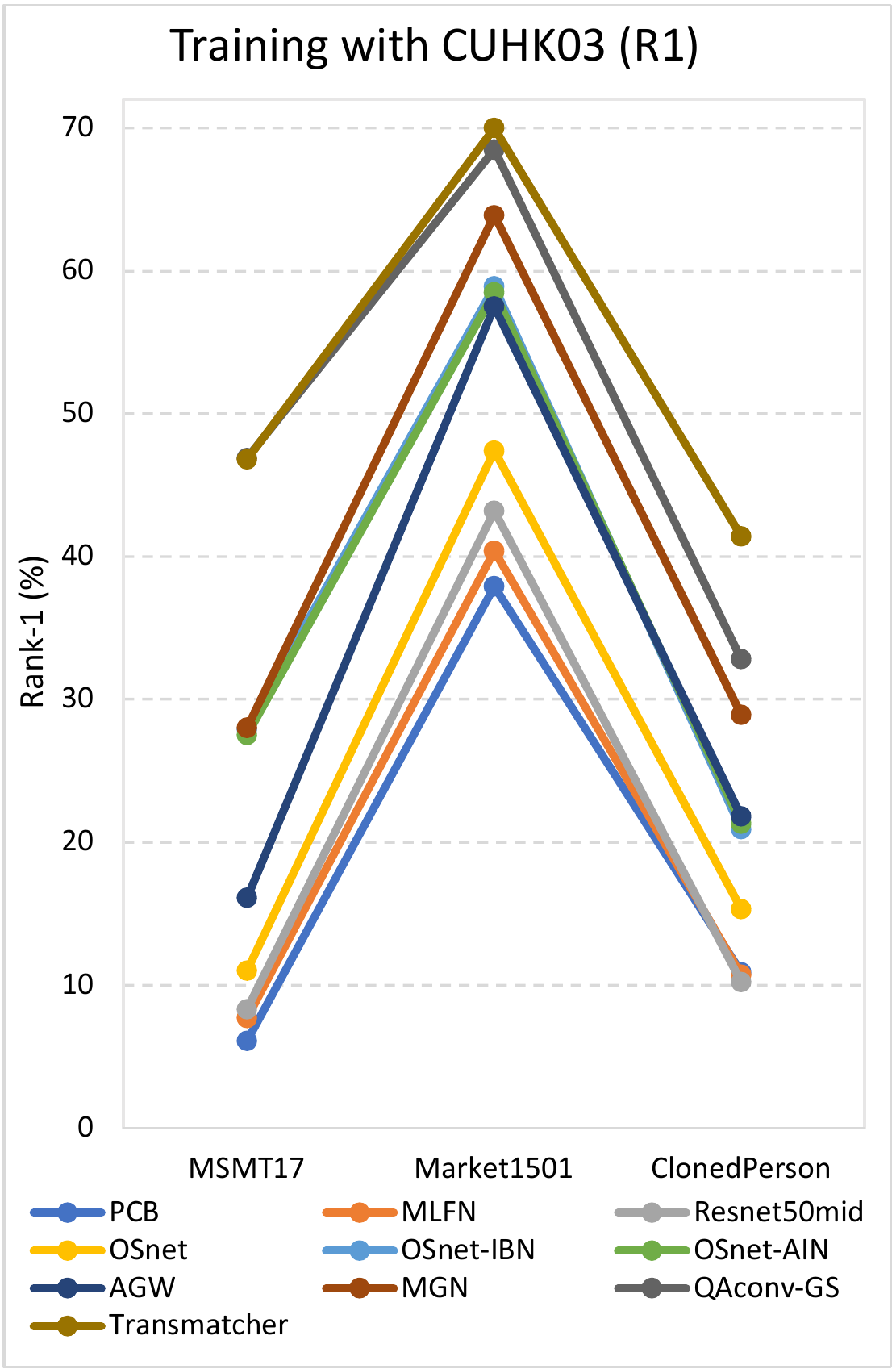}
  \caption{}\label{fig:cuhk-r1}
  \end{subfigure}
  \begin{subfigure}[b]{0.203\textwidth}
  \includegraphics[width=\textwidth]{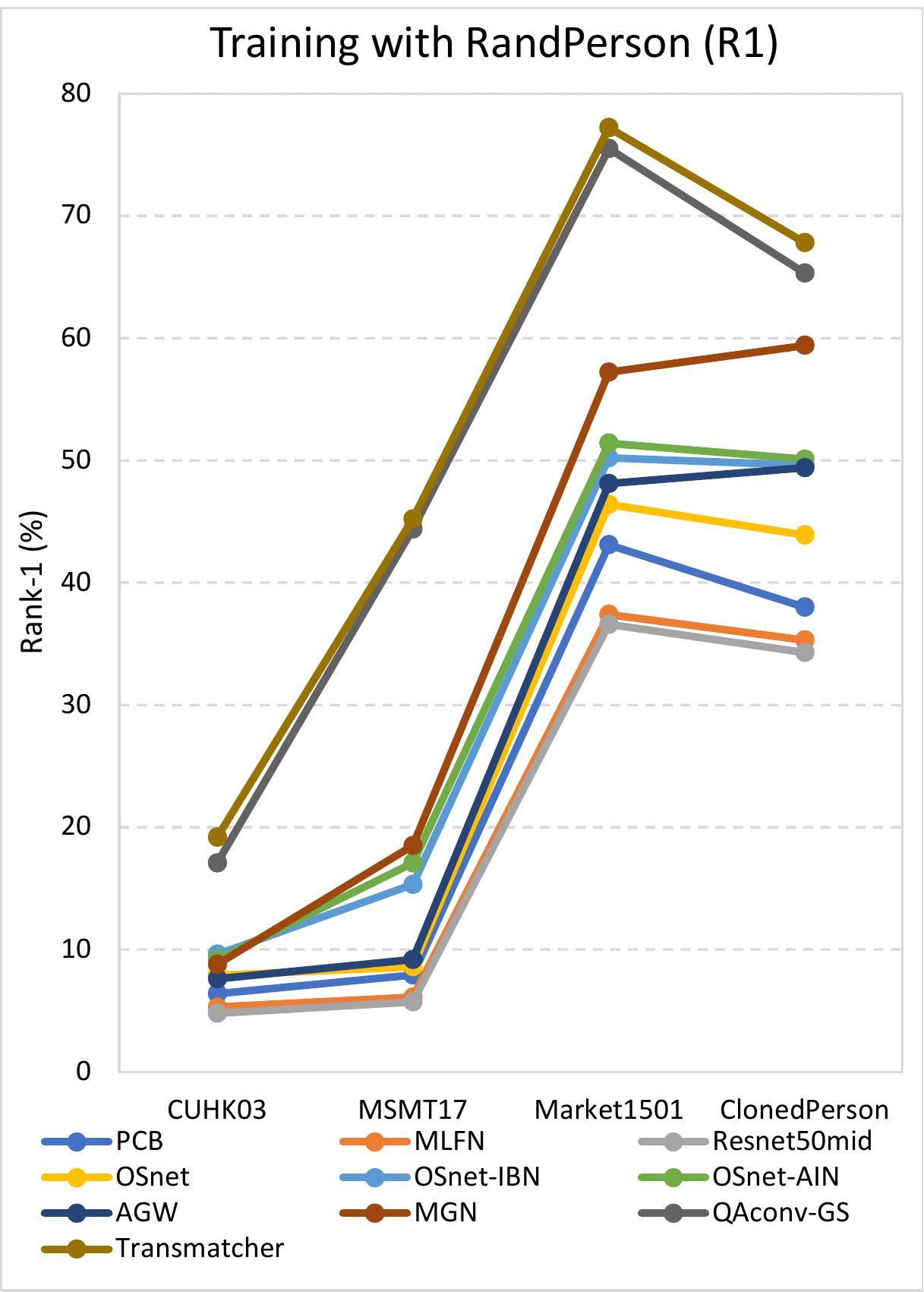}
  \caption{}\label{fig:rand-r1}
  \end{subfigure}
  \begin{subfigure}[b]{0.205\textwidth}
  \includegraphics[width=\textwidth]{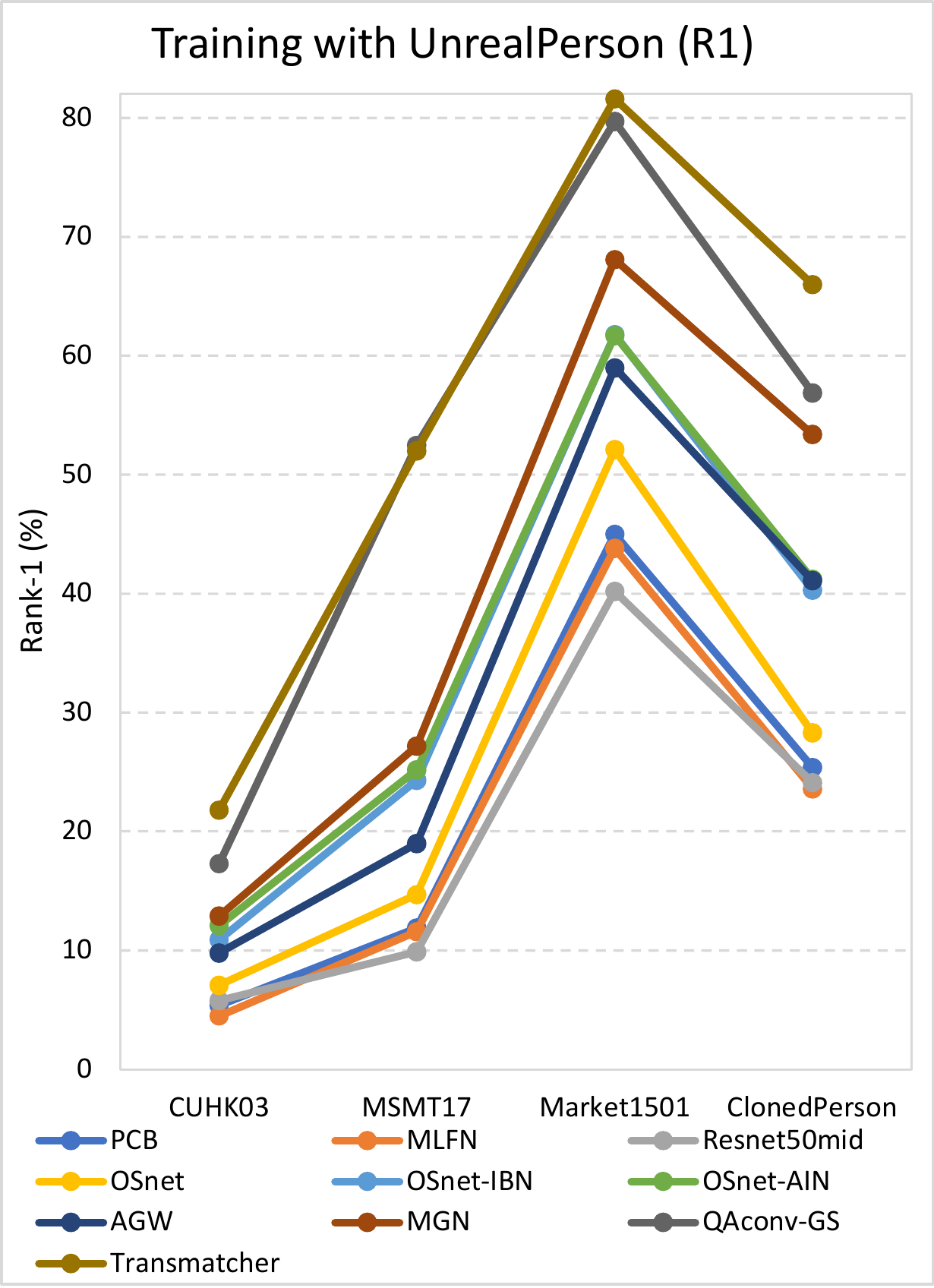}
  \caption{}\label{fig:unreal-r1}
  \end{subfigure}
  \begin{subfigure}[b]{0.19\textwidth}
  \includegraphics[width=\textwidth]{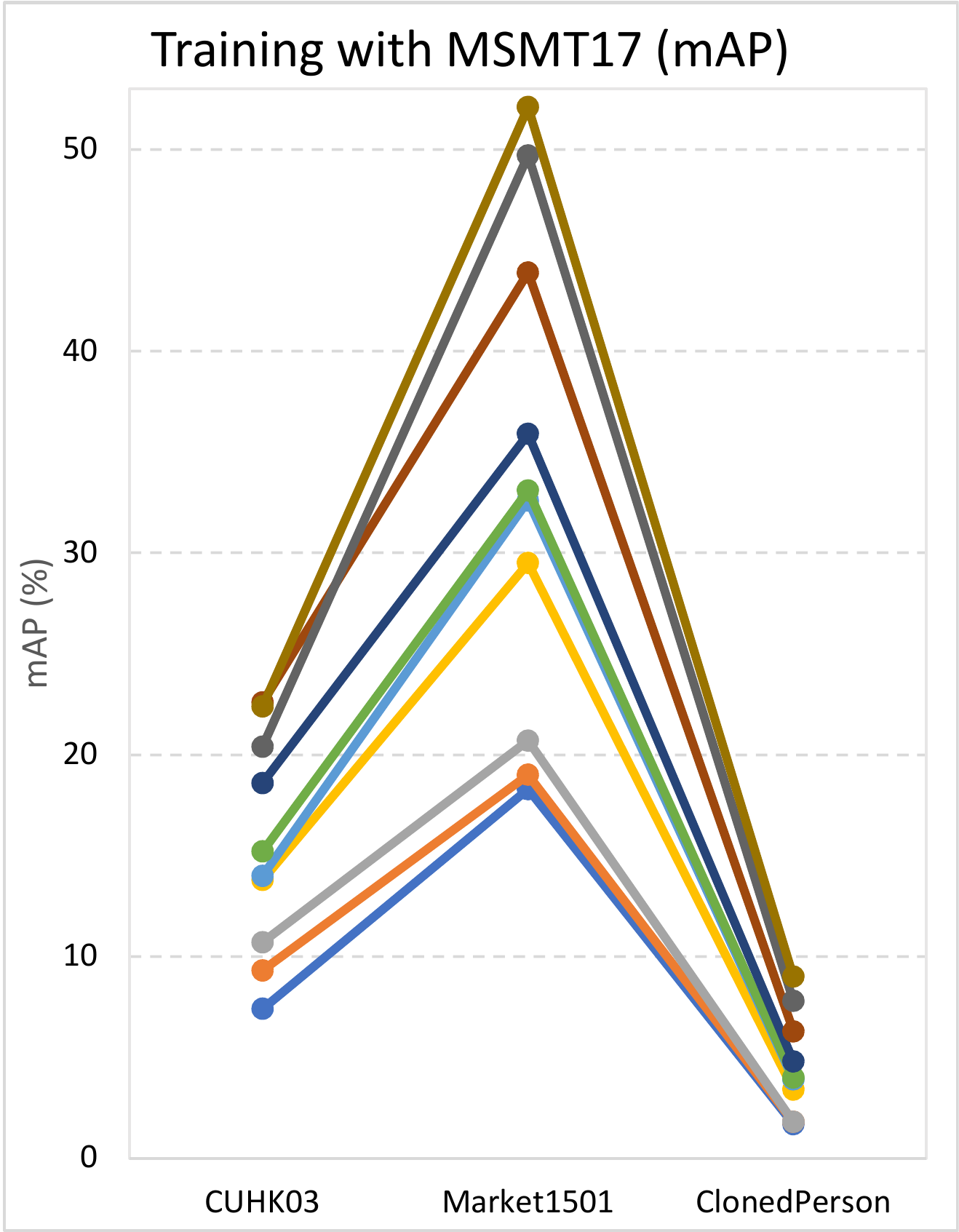}
  \caption{}\label{fig:msmt-map}
  \end{subfigure}
  \begin{subfigure}[b]{0.184\textwidth}
  \includegraphics[width=\textwidth]{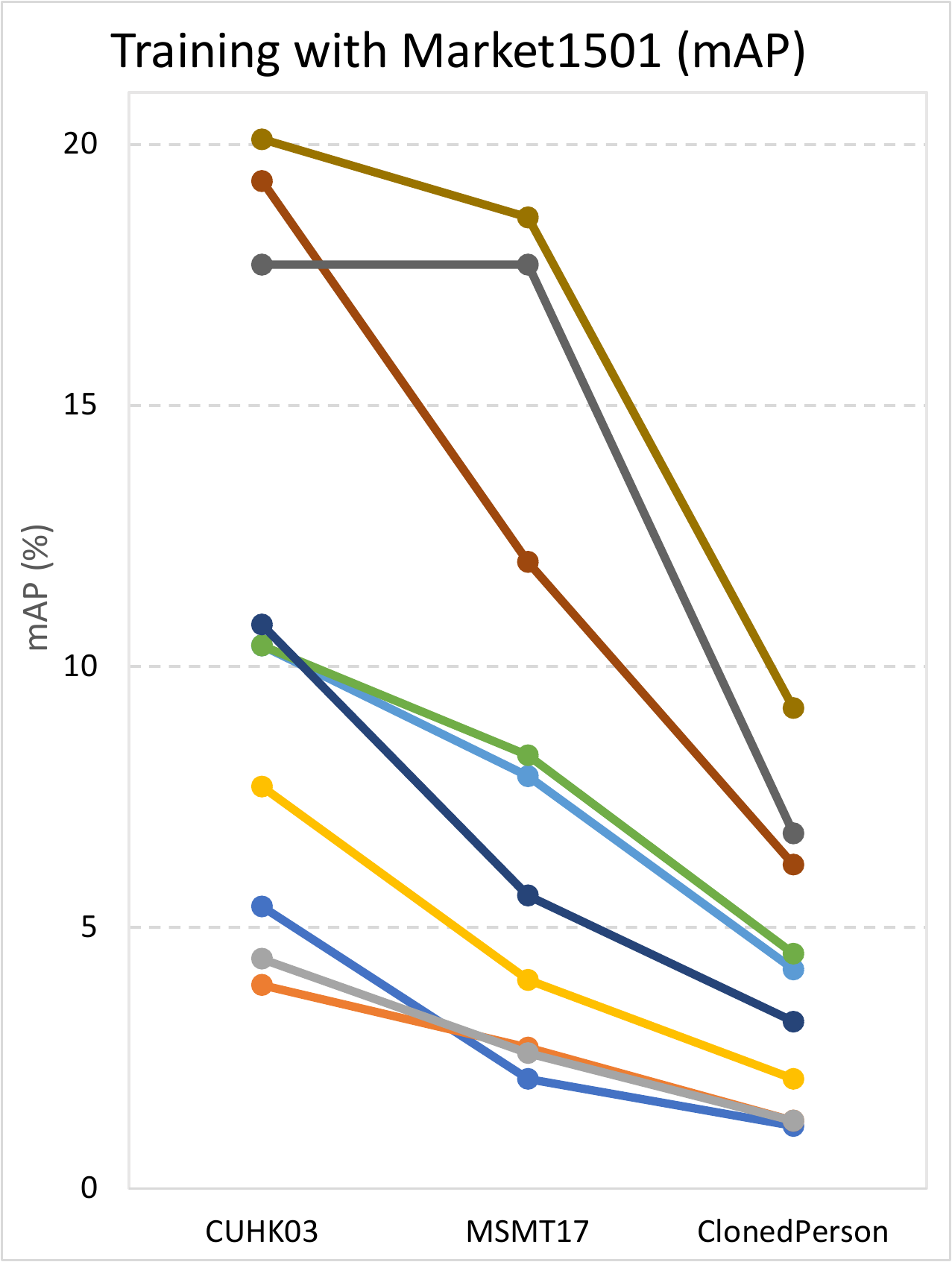}
  \caption{}\label{fig:mart-map}
  \end{subfigure}
  \begin{subfigure}[b]{0.184\textwidth}
  \includegraphics[width=\textwidth]{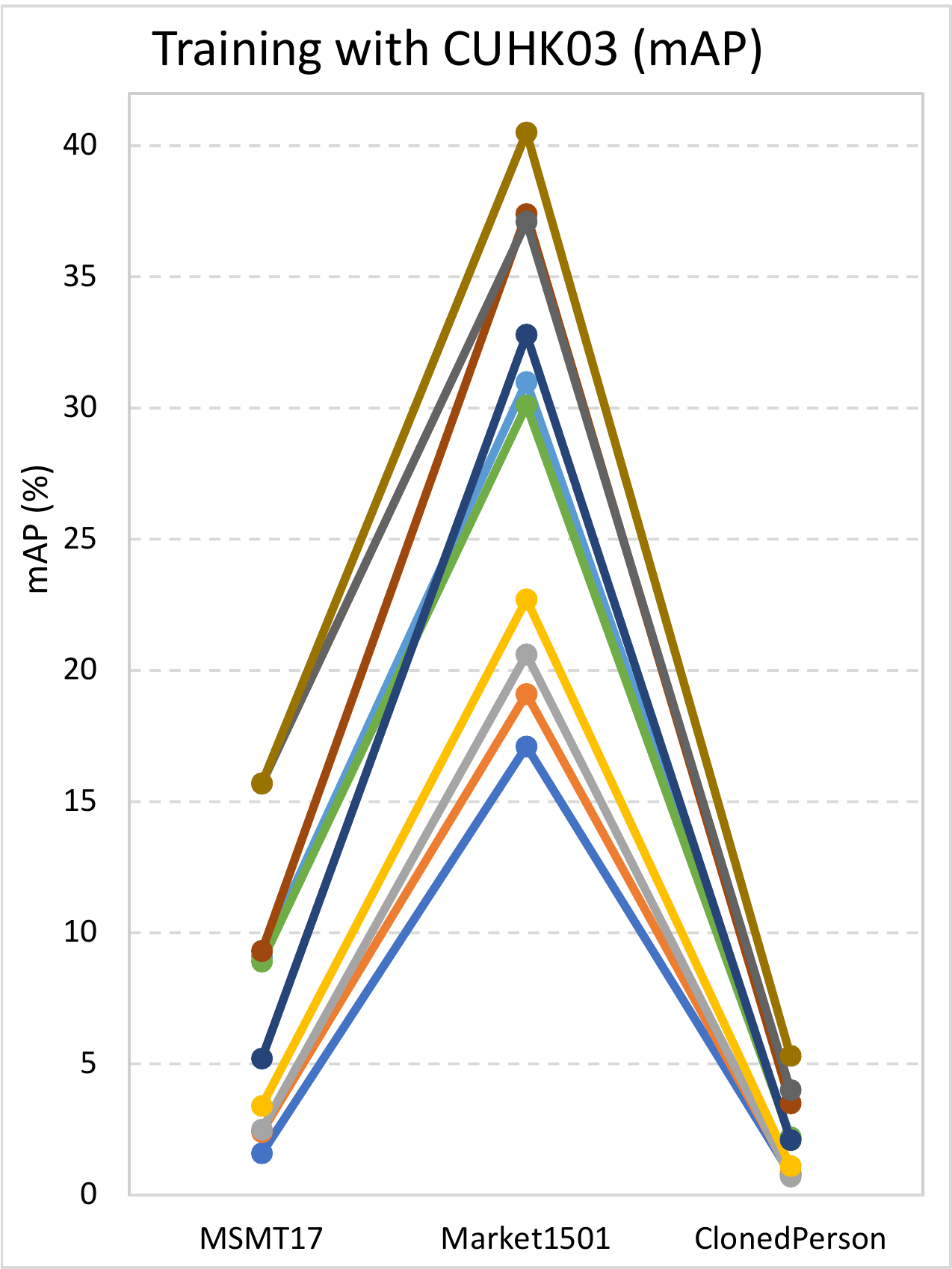}
  \caption{}\label{fig:cuhk-map}
  \end{subfigure}
  \begin{subfigure}[b]{0.2032\textwidth}
  \includegraphics[width=\textwidth]{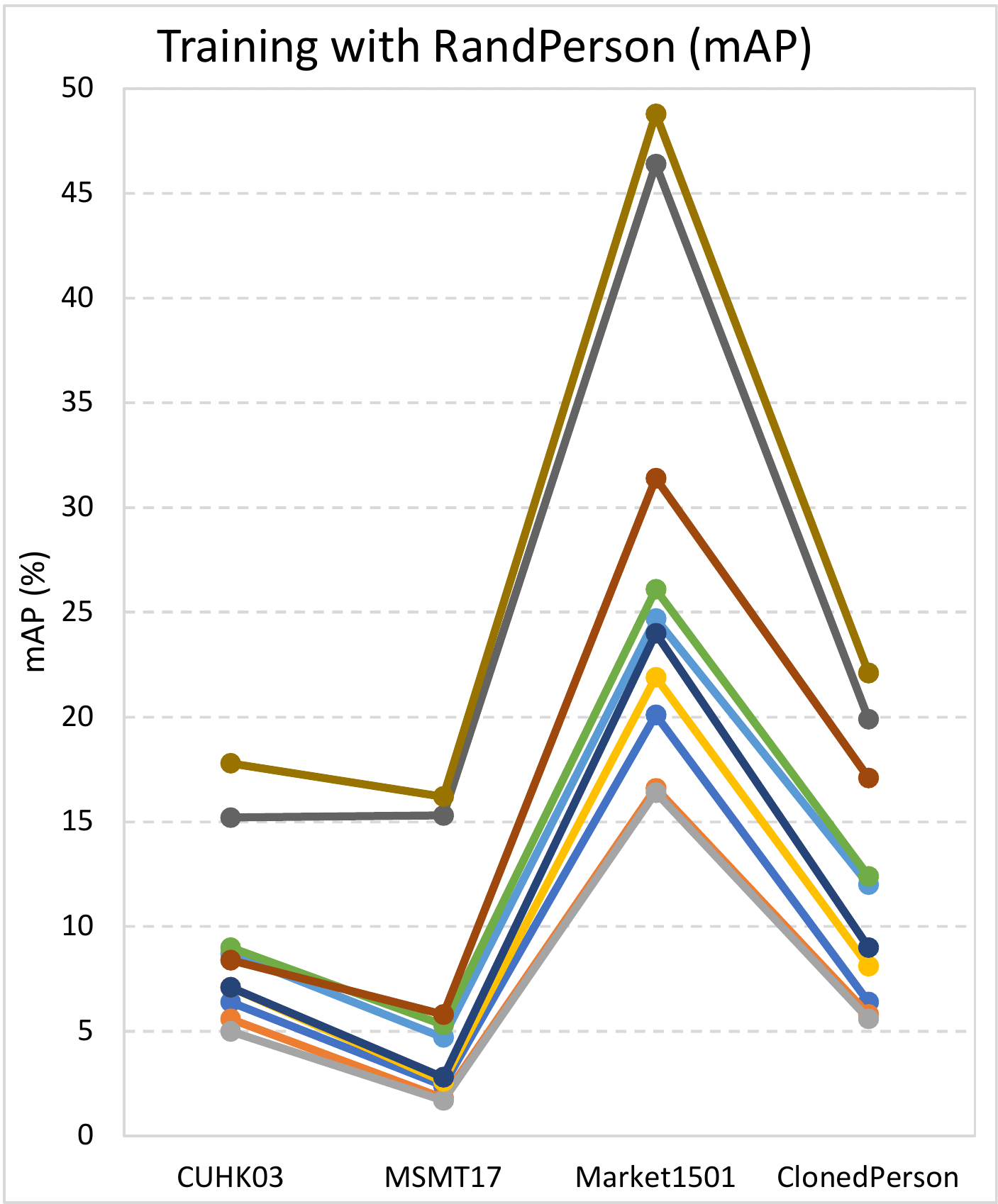}
  \caption{}\label{fig:rand-map}
  \end{subfigure}
  \begin{subfigure}[b]{0.203\textwidth}
  \includegraphics[width=\textwidth]{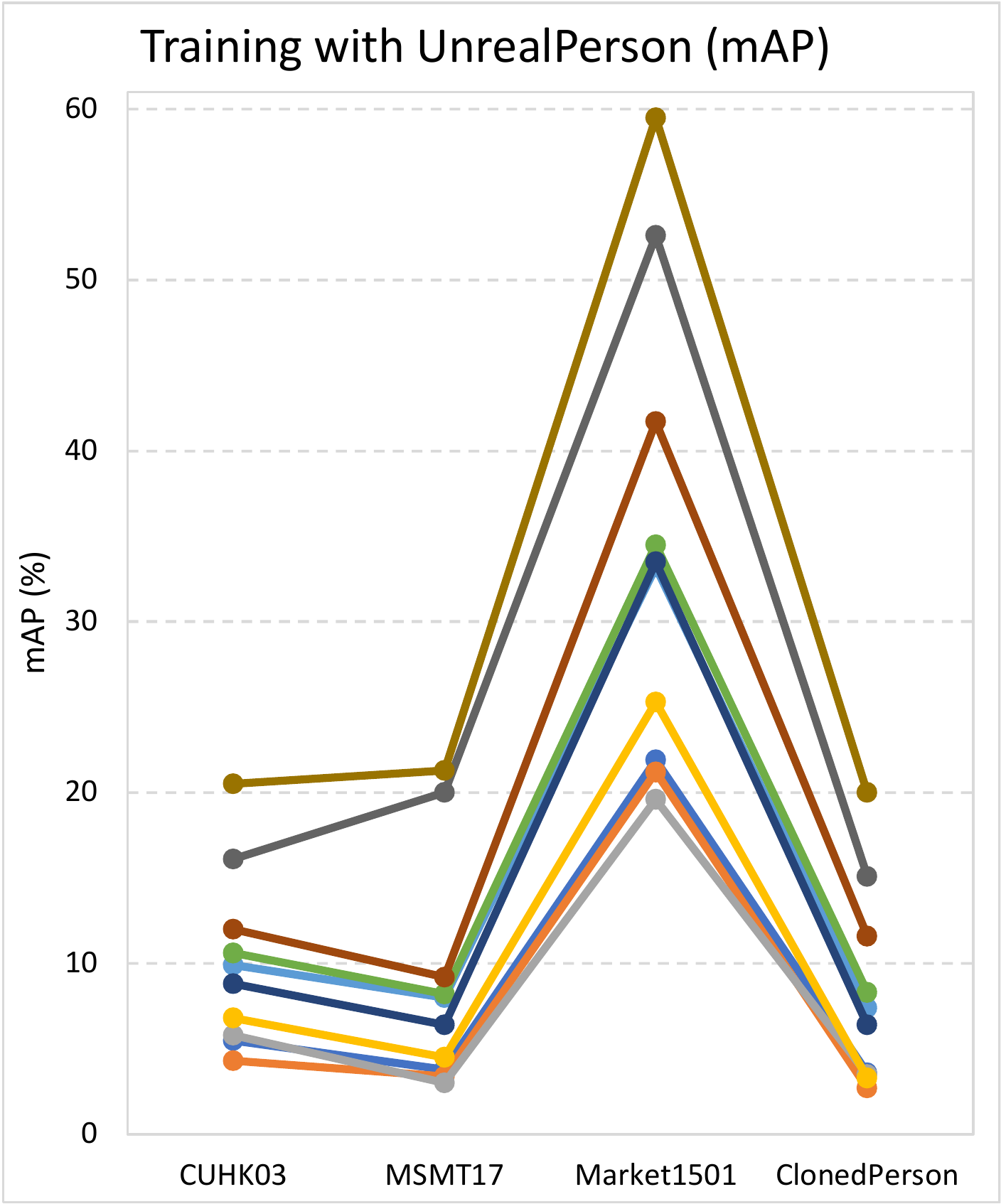}
  \caption{}\label{fig:unreal-map}
  \end{subfigure} 
\caption{Performance plots of evaluated methods with testing datasets.}
\label{fig:long}
\end{figure*}

\subsection{Results and Qualitative Analysis}

The benchmarking results of the ten algorithms with different training datasets are shown in Tables \ref{tlb:cross-msmt}-\ref{tlb:cross-unreal}.
From the tables we can see that the performance relationship of the evaluated methods on synthetic dataset is maintained almost the same with evaluation on real-world datasets. Specifically, the ten methods can be roughly divided into three groups. The top group contains TransMatcher, QAConv-GS, and MGN, with TransMatcher being the best algorithm, followed by QAConv-GS and MGN on average. The middle group contains OSNet-AIN, OSNet-IBN, and AGW, while the last group includes OSNet, ResNet50-mid, MLFN, and PCB.

\begin{table*}
\centering
\begin{tabular}{|l@{}|c c|c c|c c|c c|c c|}
\hline
Training Dataset& \multicolumn{2}{c|}{MSMT17} & \multicolumn{2}{c|}{Market-1501}& \multicolumn{2}{c|}{CUHK03}& \multicolumn{2}{c|}{RandPerson}& \multicolumn{2}{c|}{UnrealPerson} \\ 

Kendall's $\tau$ & $\tau_{R1}$ & $\tau_{mAP}$& $\tau_{R1}$ & $\tau_{mAP}$& $\tau_{R1}$ & $\tau_{mAP}$& $\tau_{R1}$ & $\tau_{mAP}$& $\tau_{R1}$ & $\tau_{mAP}$ \\ \hline
(MSMT17, Market-1501)   & -- & -- & -- & -- & 0.96 & 0.85 & 1.00 & 1.00 & 0.91 & 0.96 \\
(MSMT17, CUHK03)       & -- & -- & 0.76 & 0.72 & -- & -- & 0.82 & 0.90 & 0.87 & 0.91 \\ 
(Market-1501, CUHK03)   & 0.82 & 0.91 & -- & -- & -- & -- & 0.82 & 0.90 & 0.87 & 0.87 \\ \hline
(MSMT17, ClonedPerson) & -- & -- & 0.87 & 0.99 & 0.69 & 0.87 & 1.00 & 1.00& 0.87 & 0.87 \\ 
(Market-1501, ClonedPerson)    & 0.82 & 0.99 & -- & -- & 0.73 & 0.75 & 1.00 & 1.00 & 0.87 & 0.82 \\ 
(CUHK03, ClonedPerson)        & 0.91 & 0.90 & 0.90 & 0.75 & -- & -- & 0.82 & 0.90 & 0.91 & 0.87 \\ 
\hline\hline
(Real, Real)         & \multicolumn{4}{l}{Mean $\tau_{R1}$: 0.8701 $\pm$ 0.070 } & \multicolumn{6}{|l|}{Mean $\tau_{mAP}$: 0.8906 $\pm$ 0.073}  \\   
(Real, Synthetic)    & \multicolumn{4}{l}{Mean $\tau_{R1}$: 0.8657 $\pm$ 0.088 } & \multicolumn{6}{|l|}{Mean $\tau_{mAP}$: 0.8921 $\pm$ 0.086}   \\  \hline 
KS test with Rank-1   & \multicolumn{4}{c}{$D_{n,m}$=0.17, p-value=0.9941} & \multicolumn{2}{|c}{KS test with mAP} & \multicolumn{4}{|c|}{$D_{n,m}$=0.25, p-value=0.8506 }    \\  \hline
\end{tabular}
\caption{The Kendall's $\tau$ values and KS test results for pairwise ranking analysis.}
\label{tlb:kenl-5task}
\end{table*}


Furthermore, for a better and more clear observation, in Fig. \ref{fig:long} we plot the Rank-1 and mAP performance curves of each algorithm across different datasets. In these plots, if one line crosses the other lines, it indicates that the performance relationship is changed between the two connected datasets. On the contrary, if there is no crossing line, it means that the performance relationships are maintained the same between different testing datasets. From Fig. \ref{fig:long} it can be clearly observed that the evaluated algorithms basically keep the relationships from real-world datasets to the synthetic ClonedPerson dataset. When there are changes, most of the time the differences in performance are small where perturbations happen. Therefore, from the qualitative analysis it appears that comparing algorithms on the synthetic dataset ClonedPerson gives a strong agreement to comparing them on real-world datasets.


\subsection{Quantitative Analysis} 

More formally, the quantitative results of the Kendall's $\tau$ values and KS test for pairwise ranking analysis are shown in Table \ref{tlb:kenl-5task}. 
For the upper half of the table, it shows individual Kendall's $\tau$ value with a fixed training dataset in column, and a pair of target test datasets in row. Furthermore, the Kendall's $\tau$ values are divided into two groups (row blocks with and without ClonedPerson). Looking at these Kendall's $\tau$ values individually, it appears that most of them are high (>0.8) between the synthetic dataset ClonedPerson and other real-world datasets. Especially, when trained on RandPerson, the Kendall's $\tau$ values for the two pairs of (Market-1501, ClonedPerson) and (MSMT17, ClonedPerson) achieve the maximum value 1, which means that the rankings of the ten algorithms are completely the same on the three testing datasets involved. This can also be observed from Table \ref{tlb:cross-rand}. Therefore, this example gives a perfect agreement for ranking algorithms on Market-1501, MSMT17, and ClonedPerson when trained on RandPerson.

However, with some not large enough Kendall's $\tau$ values (e.g. 0.69), it is still not easy to judge without seeing the statistics of the other group. Therefore, more statistically, we calculate the mean and standard variance of Kendall's $\tau$ for the two groups, that is, pairs of (real-world data, real-world data) and (real-world data, synthetic data), as shown in the first two rows of the lower half of Table \ref{tlb:kenl-5task}. Now it is more clear that the average correlation coefficients are very close to each other in the two groups, with an average of 0.8657 on pairs (real-world data, synthetic data), compared to that of 0.8701 on pairs (real-world data, real-world data) for the Rank-1 measurement. Besides, the ranking correlations of the mAP measurement for (real-world data, synthetic data) is 0.8921 on average, which is also very close to 0.8906 for (real-world data, real-world data). Further considering the small standard variances, it appears that they are from two very similar normal distributions.

However, we do not have any prior assumption on the distribution type of the Kendall's $\tau$ values. Therefore, more formally, we performed the non-parametric two-sample Kolmogorov-Smirnov test to verify whether the correlation coefficients in the two groups (real-world data, real-world data) and (real-world data, synthetic data) are from identical distribution or not. The results are shown in the last row of Table \ref{tlb:kenl-5task}. Based on the Rank-1 measurement, the KS statistic for the two distributions is 0.17, and the p-value is 0.9941. Given $n$ = 9 and $m$ = 12 for the number of samples in the two groups, and the significance level $\alpha$ = 0.05, according to Eq. (\ref{eq:D-thr}), the accepting/rejecting threshold is 0.5989. Since 0.17 is significantly less than 0.5989, we accept the null hypothesis, that is, the distributions of correlation coefficients from the two groups are identical. This judgement is the same if we consider the p-value, because 0.9941 is even more significantly larger than the significance level $\alpha$ = 0.05. 
As for the mAP measurement, since the KS statistic is 0.25 and the p-value is 0.8506, we can easily draw the same conclusion. In summary, the distribution of the correlation coefficients of algorithm ranking results between the synthetic dataset ClonedPerson and other real-world datasets is identical to that only between real-world datasets. Therefore, it can be concluded that the synthetic dataset ClonedPerson can be reliably used to benchmark generalizable person re-identification algorithms, with no statistical difference to real-world datasets.

\subsection{Discussions}

The reason why testing algorithms on ClonedPerson achieves a strong agreement to that on real-world datasets is probably because in ClonedPerson \cite{Wang-2022-Clonedperson} the authors proposed an effective method to clone the whole outfits from real-world person images, so that their created 3D characters appear quite similar to real-world person images in dress.

From Tables 1-3 the results show that the mAP values on ClonedPerson are quite small when algorithms are trained on real-world datasets. However, our qualitative and quantitative analyses show that even with these small mAP values ClonedPerson is still able to distinguish and rank algorithms the same as real-world datasets. This indicates that ClonedPerson is sensitive to subtle differences of mAP performance in distinguish algorithms, and so 
every algorithm advancement will be correctly reflected. On the other hand, this indicates that ClonedPerson is a very challenging dataset for benchmarking, and given the conclusion that ClonedPerson can be reliably used for benchmarking, there is still a large space for algorithm advancement in future.

Furthermore, if we compare results of Tables 4-5 with training on synthetic datasets to Tables 1-3 with training on real-world datasets, we can observe that training on synthetic datasets achieves significantly better results on ClonedPerson than training on real-world datasets. This is consistent with \cite{wang20rand,zhang2021unrealperson,Wang-2022-Clonedperson} and further confirms their findings. Now that ClonedPerson is statistically accepted for benchmarking algorithms without difference to real-world datasets, this observation suggests that for the study of GPReID, we can completely use synthetic datasets from source training set to target testing set, with the advantages of both improved performance and completely no privacy concerns from real-world surveillance data.


\section{Conclusion}

In this paper, we studied the problem of whether synthetic datasets can be reliably used for benchmarking generalizable person re-identification. Through the designed pairwise ranking analysis method and comprehensive evaluations, we conclude that the recent large-scale synthetic dataset ClonedPerson can be reliably used to benchmark GPReID, statistically the same as real-world datasets. Furthermore, we find that for the study of GPReID, using synthetic datasets for both source training set and target testing set is in favored, with the advantages of both improved performance and completely no privacy concerns from real-world surveillance data. In future research, we could also study using synthetic datasets to benchmark other person re-identification tasks, for example, domain adaptation and unsupervised learning. Besides, the study in this paper might also inspire future designs of synthetic datasets.

{\small
\bibliographystyle{ieee}
\bibliography{reid}
}

\end{document}